\documentclass[a4paper,fleqn]{cas-dc}
\usepackage[numbers]{natbib}

\usepackage{algorithm}
\usepackage{algpseudocode}
\usepackage{amsmath}
\usepackage{geometry}
\usepackage{algorithmicx}
\usepackage{algpseudocode}

\def\tsc#1{\csdef{#1}{\textsc{\lowercase{#1}}\xspace}}
\tsc{WGM}
\tsc{QE}
\tsc{EP}
\tsc{PMS}
\tsc{BEC}
\tsc{DE}

\begin{document}
\let\WriteBookmarks\relax
\def\floatpagepagefraction{1}
\def\textpagefraction{.001}

\shorttitle{Leveraging social media news}

\shortauthors{Wei Xue et~al.}

\title [mode = title]{Stereo Matching with Cost Volume based Sparse Disparity Propagation}                      


%
\author[1,2]{Wei Xue}[style=chinese]





\address[1]{College of Big Data and Internet, Shenzhen Technology University, Shenzhen, China}
\address[2]{College of Computer Science and Software Engineering, Shenzhen University, Shenzhen, China}

\author[1]{xiaojiang Peng}[style=chinese]
\cormark[1]


\cortext[cor1]{Corresponding author }
\nonumnote{pengxiaojiang@sztu.edu.cn (X. Peng)}
\nonumnote{URL: https://pengxj.github.io (X. Peng)}
\nonumnote{KimPhillipp@outlook.com (W. Xue)}




\begin{abstract}
Stereo matching is crucial for binocular stereo vision. Existing methods mainly focus on simple disparity map fusion to improve stereo matching, which require multiple dense or sparse disparity maps. In this paper, we propose a simple yet novel scheme, termed as feature disparity propagation, to improve general stereo matching based on matching cost volume and sparse matching feature points. Specifically, our scheme first calculates a reliable sparse disparity map by local feature matching, and then refines the disparity map by propagating reliable disparities to neighbouring pixels in the matching cost domain. In addition, considering the gradient and multi-scale information of local disparity regions, we present a $\rho$-Census cost measure based on the well-known AD-Census, which guarantees the robustness of cost volume even without the cost aggregation step. Extensive experiments on Middlebury stereo benchmark V3 demonstrate that our scheme achieves promising performance comparable to state-of-the-art methods.

\end{abstract}



\begin{keywords}
Disparity Fusion \sep Stereo Matching \sep Feature Matching \sep 
Ad-Census\sep 
Local Expansion
\end{keywords}

\maketitle

\section{Introduction}
Stereo matching is an important part of inferring depth by extracting corresponding pixels in image pairs. The technique is extensively employed in industrial production (mapping system), automobile (self-driving system), aerospace (planetary probe landing system, remote sensing system), etc., contributing to machine capability of environment perception. 
For better depth estimation capability, several proposals have been presented. 
However,is rarely stereo matching associated with feature matching, which is similar to it.

Feature matching is to match the points in the image pair that have local feature invariance (i.e., features points) and obtain the correspondences between the feature points in different images. This correspondence relationship is very important for tasks that require spatial geometric relationships, such as SLAM(Simultaneous Localization and Mapping) and SfM(Structure-from-Motion). After the efforts of the researcher, the feature matching method is constantly being improved: from the classic traditional methods that use SIFT \cite{ref2}, SURF \cite{ref8}, and other feature descriptors for matching (such as FLANN \cite{ref7}) to the latest SuperGlue \cite{ref3} and other end-to-end methods based on neural network. At the same time, there are also methods to calculate features based on trained networks, such as SuperPoint \cite{ref6}. These allow us to obtain more feature points from the image, more feature point pairs that are successfully matched, and the matching accuracy rate is getting higher. However, compared with the disparity map obtained by the stereo matching algorithm, the disparity map obtained by the feature matching is still too sparse, but the corresponding relationship between these successfully matched feature points is very accurate, robust, and reliable.


At present, there are Binocular stereo methods and Multi-view stereo methods. In this paper, we only study Binocular stereo matching methods, so stereo matching specifically refers to Binocular stereo matching in this paper. According to the research of Scharstein and Szeliski \cite{ref1}, most stereo matching algorithms’ process is summarized into the following four steps: matching cost computation, cost aggregation, disparity computation and disparity refinement. Its essence is to find pixel point pairs corresponding to the same object in the real world in two images with different perspectives. In order to improve the performance of matching, researchers have added different constraints, such as colour constraints, gradient constraints, and smoothness constraints. After many years of development, the stereo matching algorithm has been continuously optimized, but there are still a considerable number of disparity values in the disparity map that are not accurate and reliable.

In this letter, a method of using feature matching is proposed to optimize the stereo matching cost volume with a reliable and accurate sparse disparity map as a priori information, to explore solutions to the above problems. Therefore, a feature disparity propagation method in cost domain is presented. Stereo matching and feature matching have great similarities, indeed we can attempt to combine feature matching algorithms to optimize the stereo matching process, but there are currently seldom researches.
The key to stereo matching is to find the correspondence point pairs of the left and right view images. The same is true for feature matching, but the latter’s matching standard is more rigid, so only fewer points can become features and get correspondence feature points in another image. Through Equ.\ref{eq1}, we can also use feature matching to get a sparse disparity map. After many experiments comparing ground truth, it is found that although the matched relationship obtained through feature matching is small, it is very reliable and robust.
\begin{equation}
d_{p}=u_{p}-u_{q} 
\label{eq1} 
\end{equation}
Where $d_p$ represents the disparity of pixel $p$. $p$, $q$ are the correspondence point pair in images. $u_p$ and $u_q$ respectively represent the abscissa of points in the image domain.

Finally, the result disparity map is generated by LocalExp \cite{ref9} as an optimizer based on slanted support
windows \cite{ref4} and graph cuts \cite{MinCutAndMaxFlow}.  It allows the object surface of result disparity map to avoid front-parallel bias \cite{ref4}, \cite{ref5}. The concept of slanted support windows was pioneered by Blyer et al. \cite{ref4} in 2011, by over-parameterizing the disparity value.

\begin{equation}
d_{p}=a_{p} u+b_{p} v+c_{p} 
\label{eq2} 
\end{equation}

Estimates disparity plane label ($a_p$, $b_p$, $c_p$) for each pixel p in the image domain ($u$, $v$) instead of directly estimating $d_p$. The LocalExp method proposes local expansion moves based on graph cuts, uses many $\alpha$-expansions in small grid regions. Based on this, LocalExp method allows better disparity labels to be propagated to neighbouring pixels. Through this process, the good disparity obtained from the sparse disparity map propagate with more possibility to the neighbouring pixels that may be on the same plane.

In this work, we tried to combine feature matching and stereo matching. The results of a series of reliable experiments show that the robust, reliable but sparse disparity map obtained by feature matching can successfully optimize stereo matching and produce a better disparity map.

In the experiments of this letter, we use the classic SIFT descriptor proposed by Lowe \cite{ref2} and KNN (K-Nearest-Neighbor) feature matching \cite{KNN} to get the sparse disparity map. Actually, there are more options, such SuperGlue feature matching method proposed by Sarline et al. \cite{ref3} and the SuperPoint feature description method proposed by DeTone et al. \cite{ref6}. Our stereo matching cost measure algorithm is based on AD-Census method \cite{Ad_censusPaper}, because it can efficiently employ local information. Intel RealSense D400 \cite{ref14}, \cite{RealSense} is based on Ad-Census to implement stereo correspondence. In the tests, we use Middlebury Stereo Dataset \cite{ref10}, \cite{ref11}, \cite{ref12}, \cite{ref13}. All methods are implemented using OpenCV (the Open Computer Vision) library \cite{ref22} based on C++. Our code will be available at GitHub (https://github.com/KimPhillippCavendish).

\section{Related Work}
\subsection{Stereo Matching}
It employs a rectified image pair captured from two viewpoints, and purposes inferring the depth information in the form of disparity. A classical four-step stereo matching algorithm framework \cite{ref1} has been summarized in 2002.

\subsubsection{Local Stereo Matching}
Local methods compute the disparity for each pixel independently. In order to achieve more robust matching cost, the cost aggregate from a local region (Aka. Aggregation window). Then the disparity with the minimal cost will be selected using Winner-Take-All method. To optimize the disparity results, methods such as sub-pixel interpolation and plane fitting are implemented. There are also optimizers such as graph cuts \cite{AlphaExpansionInGC} and belief propagation \cite{BP1}, \cite{BP2} similar to the global matching method, where they optimize the disparity map by minimizing a global energy function. 

\subsubsection{Global Stereo Matching}
Generally, global methods are more precise and smooth than local methods, but far away from being real-time. During the global method, not only the photo-consistency $\phi(d_p)$ between matching pixels is calculated, but also the smoothness $\psi_{p q}(d_{p}, d_{q})$ between every pixel $p$ and its neighbour pixels $q$ is measured.
Conventional MRF stereo methods \cite{AlphaExpansionInGC}, \cite{EnergyOptiUsingGC}, \cite{DefenseOf3D}, \cite{GlobalLeveragedSparse} are implemented by constructing an energy formulation \ref{EnergyFunction} \cite{ref9} based on a pairwise MRF:

\begin{equation}
E(f)=\sum_{p \in \Omega} \phi_{p}\left(d_{p}\right)+\lambda \sum_{(p, q) \in \mathcal{N}} \psi_{p q}\left(d_{p}, d_{q}\right)
\label{EnergyFunction}
\end{equation}

The first term $\phi(d_p)$ is referred to as the \textit{data term} or \textit{unary term}. The image domain $\Omega$ contains every pixel $p$ in the input left and right images. The second term $\psi_{p q}(d_{p}$, $d_{q})$ is named \textit{smoothness term} or \textit{pairwise term}, which penalizes unsmoothness of the disparity map (namely, discontinuity of disparities between neighbouring pixel pairs $(p,q)\in \mathcal{N}$ \cite{ref9}).

\subsubsection{Cost Volume}
In the conventional stereo vision pipeline, matching cost computation and cost volume construction is an essential component of the whole process. At present, the state-of-the-art methods are all based on the cost volume \cite{CFNet}. The traditional cost volume $CM$ is represented as a 3D array of size \textit{Height $\times$ Width $\times$ Number of disparity} of the input images, and each element of the array stores a cost value.

$CM(x,y,d)$ mentioned in Section 4.1 refers to the matching cost of pixel $(x,y)$ at disparity $d$, namely the cost of the correspondence between pixel $(x, y)$ and pixel $(x-d, y)$.

As summarized in \cite{CFNet}, the methods for generating cost volume fall into two categories. The first category uses a full correlation to create a 3D cost volume based on a single-feature measure. The methods of the second category generally generate a multi-feature 4D cost volume using concatenation \cite{concatenation} or group-wise correlation \cite{correlation}. 

The first-category methods are typically more efficient, but some feature profiles could not be exploited due to the single matching measure. The second-category methods, which including CFNet \cite{CFNet}, Dispnet \cite{Dispnet}, and AANet \cite{AANet}, sacrifice more time and computational resources while providing better results. 

Our proposal for updating cost volume falls into the first category, but uses a reliable sparse map to improve the initial cost volume. In this proposal, the cost volume is improved while the computational complexity and memory consumption is lower than the methods in the second category.

In our work, an improved AD-Census (aka. $\rho$-Census, which is illustrated in section \ref{rho-Census}) is proposed for constructing the initial cost volume.

\subsubsection{AD-Census}
AD-Census \cite{Ad_censusPaper} is a manually designed GPU-based stereo corresponding measure proposed by X. Mei et al, which is also implemented in Intel RealSense RS400 series \cite{RealSense}. Ad-Census measure is able to extract the local distinction features of the left and right view graphs nicely. And since AD-Census measure is designed to achieve a high parallelism, both matching accuracy and processing efficiency are well-balanced. 

As its name indicates, AD-Census matching cost consists of two parts, AD (Absolute Difference) cost and Census \cite{census} cost. Given a pixel $p = (u_p, v_p)$ defined on the reference image domain $I^{Left}$, the correspond pixel $q = (u_q,v_q)$ defined on the match image domain $I^{Right}$, and a disparity level $d$. There is obviously that $u_p - d = u_q$ and $p = q - (d, 0)$. AD cost value $C_{AD}(p,d)$ is defined as the average intensity difference of pixel $p$ and correspond pixel $q$ 
in RGB channels \cite{census}:

\begin{equation}
C_{A D}(p, d)=\frac{1}{3} \sum_{i=R, G, B}\left|I_{i}^{L e f t}(p)-I_{i}^{R i g h t}(q)\right|
\label{AD}
\end{equation}

Census cost value $C_{census}(p,d)$'s definition is the Hamming distance of the two bit strings that stand for pixel $p$ and $q$ \cite{Ad_censusPaper}. After $C_{AD}(p,d)$ and $C_{census}(p,d)$ have been normalized by robust function $\theta$:
\begin{equation}
\theta(c, \lambda)=1-\exp \left(-\frac{c}{\lambda}\right),
\label{functionTheta}
\end{equation}
the AD-Census cost value $C_{ad-census}$ is computed as follows:
\begin{equation}
\begin{aligned}
C({p}, d)=& \theta\left(C_{\text {census }}({p}, d), \lambda_{\text {census }}\right)+\\
& \theta\left(C_{A D}({p}, d), \lambda_{A D}\right)
\end{aligned}
\label{adcensusComput}
\end{equation}

On the basis of the classical AD-Census measure, we fused the gradient information to calculate the first term of Equation \ref{adcensusComput}, and used multi-scale Census cost instead of the traditional Census cost in the second term. More details are covered in section \ref{rho-Census}.

\subsection{Multi Disparity Fusion}
For improving the performance of result disparity map, a number of approaches have been proposed to estimate result disparity map from the fusion of multiple disparity maps. 
The disparity maps can be acquired from different subsystems for depth measurement, such as Stereo-ToF fusion \cite{ProbabilisticFusion}, Lidar-Stereo fusion \cite{LiDarStereo}, \cite{RealTimeProb}. It can also be generated from the same depth measurement system driven by various algorithms, such as different stereo matching methods in a stereo vision system (namely, Stereo-stereo fusion \cite{DeepStereoFusion}).

Fusing the result disparity maps generated from various depth measurement subsystems allows compensating the weaknesses of the subsystems to a certain extent.
The disparity data acquired by stereo pair is denser (i.e., higher resolution), while the disparity data obtained by ToF or structured light is more accurate. 
In addition, ToF systems and structured light systems are more expensive and it is more costly to upgrade them.

The multi disparity fusion based methods improve the reliability of disparity data obtained from a single method, while maintaining the resolution and denseness of disparity maps. 
Presently, beside the MAP- MRF framework based methods like probabilistic fusion \cite{ProbabilisticFusion}, there are more methods driven by Deep Learning to fuse disparity data: UDFNET \cite{UDFNET}, DSF (Deep Stereo Fusion) \cite{DeepStereoFusion} and a CNN based disparity map fusion network \cite{LiDarStereo}.

The depth measurement subsystems in fusion based approaches can estimate disparity (equivalent to depth) maps independently.
However, their efforts (\cite{UDFNET}, \cite{DeepStereoFusion}, \cite{LiDarStereo}) mainly focus on the fusion of simple disparity maps, and the matching cost information in cost volume still underutilized. 
In Section 4, we present a disparity propagation methodology in cost volume domain. 

Additionally, the fusion of stereo vision system with other depth measurement systems requires the installation of extra hardware devices, which increases the spatial and economic burden.
In this letter, we attempt to combine feature matching algorithm with stereo matching algorithm based on the single stereo vision system. 

\subsection{Feature Matching}
Feature matching is generally performed in the following main steps \cite{ref3}:
\begin{enumerate}
\item Detecting Interest points 
\item Computing visual descriptors 
\item Matching using descriptors a Nearest Neighbour (NN) search
\item Filtering incorrect matches
\item Estimating a geometric transformation
\end{enumerate}

In steps 1 and 2, feature points with a set of feature invariants can be derived and characterized by descriptors. Then, the process of comparing different feature descriptors to find the most similar feature points is namely feature matching. By matching the feature points of two different view images, feature point pairs can be obtained, and the disparity value of the feature points can be calculated. This is an important component of the implementation of SLAM system \cite{SLAM}.

Since the criteria for extracting feature points in feature matching are more stringent than those for stereo matching, the accuracy of feature matching is better than that of stereo matching. However, the feature points are sparse, and it would be impossible to find corresponding matches for all points in image pair, which results in a sparse depth map. In this work, we try to use our proposed method to combine stereo matching and feature matching to generate a more accurate and dense disparity map.

\subsection{Continuous MRF Stereo Method: LocalExp}
We use LocalExp \cite{ref9},\cite{LEGC} proposed by T. Taniai et. as post-processing optimizer, which is a continuous MRF stereo method like \cite{PMBP}, \cite{PM-Huber}, \cite{ref4}, \cite{PMFilter}, \cite{DefenseOf3D}, \cite{GlobalStereoRecons}. It is an accurate stereo corresponding method using local expansion moves \cite{AlphaExpansionInGC} proposed based on graph cuts (GC) \cite{MinCutAndMaxFlow}, \cite{GC}.

LocalExp can either use randomized initialization like PatchMatch \cite{ref4}, assigning each pixel a random initial disparity plane, or use externally provided cost volume as an aid to generate the initial disparity plane. LocalExp differs from fusion moves-based \cite{FusionMoves4MRF} methods such as \cite{DefenseOf3D}, \cite{GlobalStereoRecons} in that LocalExp uses spatial propagation and randomization search during inference using an initial solution. Through spatial propagation, the "good" disparity labels are allowed to be propagated to pixels with appropriate positions and colours. Randomization search enables the algorithm to explore candidates other than the labels to be propagated. In our scheme, we want to use these two processes to improve the propagation of feature disparity.

\section{Proposed Method}
\begin{figure}[ht]
\centering
\includegraphics[width=0.46\textwidth]{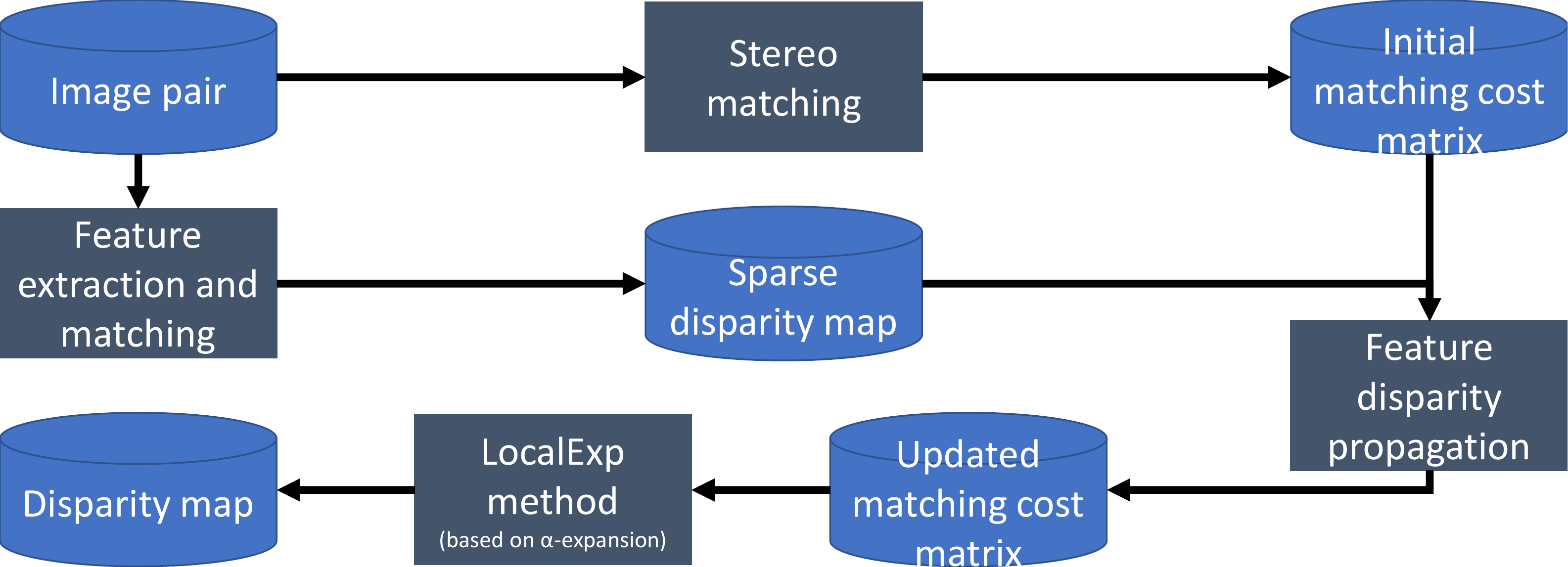}
\caption{Framework of proposed method}
\label{img1}
\end{figure}

The steps of our method are presented in Fig. \ref{img1}. Our method's input is a stereo rectified image pair taken by a binocular camera. The output of our method is an estimated dense disparity map. This section will describe the key process of our method.

In Section 3.1, we first describe how the initial matching cost matrix in this work is computed. Then in Section 3.2, we explain the procedure to obtain the Sparse disparity map. The key of this method, Feature Disparity Propagation, is designed to combine sparse and dense disparity maps, which will be introduced in Section 3.3.

\subsection{Initial Matching Cost Computation}
\label{rho-Census}
To obtain the initial matching cost, we used the improved AD-Census cost measure which has good performance in both accuracy and speed. In this step, other cost computing methods can also be used, such as MI(Mutual Information) \cite{ref20}, BT \cite{ref21}.

AD-Census cost is composed of the absolute differences and cost of Census. In the improved AD-Census measure, we used multi-scale Census cost, and inspired by \cite{ref4}, we use $\rho(p,d)$ function instead of the absolute differences (AD). Therefore, we call the improved AD-Census measure the \textit{$\rho$-Census} measure, which is one of our main contributions.

Given a pixel $p(x,y)$ with disparity value $d$, $S_{census}(p,d)$ and $\rho(p,d)$ are calculated first. $S_{census}(p,d)$ is defined as the sum of the census cost of pixel $p(x,y)$ at disparity $d$ at each scale, and image pairs at different scales are generated by different levels of Gaussian blurring. $S_{census}(p,d)$ is computed as follows:
\begin{equation} S_{census}(p,d)=\sum_{i=0,i\in N}^{n}w_i \cdot C_{census,i}(p,d) \label{eq3} \end{equation}
where different $i$ means different scales, the larger the $i$ the higher the blurring of the image pair at that scale $i$, when $i$=1 it represents the original image pair. $N$ is the scale size, indicating that the census cost computing is performed at $N$ scales. $C_{census,i}(p,d)$ represents $C_{census}(p,d)$ on scale $i$. $C_{census}(p,d)$(the census cost value of pixel $p$ at disparity $d$) is defined as the Hamming distance of the two bit strings that stand for pixel $p$ and its correspondence pixel $q(x-d,y)$ in the right image \cite{census}. In this work, we take $n$=3 and $w1$, $w2$, $w3$ to be 0.7, 0.2, 0.1 respectively.

The $\rho$ function has been applied in \cite{ref18}. The function additionally uses gradient information to better measure the similarity of pixels than AD. In this paper, a few changes have been made to the function to accommodate this work:
\begin{equation}\begin{aligned}\rho(p, q)=&(1-\alpha) \cdot \frac{\left\|I_{p}-I_{\mathrm{q}}\right\|}{3}+\\&\alpha \cdot\left\|\nabla I_{p}-\nabla I_{\mathrm{q}}\right\| \label{eq4}\end{aligned} \end{equation}
$q(x-d,y)$ is the correspondence of $p(x,y)$ in the right image. $||I_p-I_q||$ denotes the Manhattan distance of the two points $p$, $q$ in the colour space. $||\nabla I_p-\nabla I_q||$ is the sum of the gradient differences between points $p$, $q$ in two directions (i.e., horizontal direction and vertical direction). The user-defined parameter $\alpha\in[0, 1]$ adjusts the influence of colour and gradient term\cite{ref4}.
Fig. \ The $\rho$-Census cost value $C(p,d)$ is then computed as follows:
\begin{equation}\begin{aligned}
C(p, d)&=\theta\left(\rho(p, p-(d, 0)), \lambda_{\rho}\right)+\\& \theta\left(S_{census}(p, d), \lambda_{census }\right)
\label{eq5}\end{aligned}\end{equation}
where $\theta(c, \lambda)$ computed by Equ. \ref{functionTheta}, which from \cite{Ad_censusPaper} is a normalized function of the variable $c$.

Store all $C(p, d)$ in a 3D matrix $CM$ (Cost Matrix). The size of the matrix $CM$ is $H\times W\times D$. $H$ is the height of the image, $W$ is the width of the image and $D$ is the range of disparity. $CM(p,d)$ represents the matching cost of pixel $p$ at disparity $d$. Also, since the position of pixel $p$ equals $(x_p,y_p)$, $CM(p,d)$ can be represented as $CM(x_p,y_p,d)$.

\subsection{Sparse Disparity Map Construction}
The sparse disparity map is constructed using feature matching. After obtaining the matching feature point pairs in the left and right views, a sparse disparity map can be obtained by Equ.\ref{eq1}. The standards for establishing feature points are very strict, only a relatively small number of pixels can be a feature point, so the standards for feature matching are more rigid than for stereo matching. Therefore, compared to stereo matching, feature matching tends to be more correct, accurate and robust, but more sparse.

Here, we use the classical SIFT feature descriptor, KNN matching method to find matching pixel pairs. To reduce the probability of matching errors, Ratio test is also added to the process.If the ratio of $MD_0$ (the matching distance between $F_0$ and $F_1$) to $MD_1$ (the  matching distance between $F_0$ and $F_2$) is greater than a certain ratio, the match is not accepted, i.e. it is considered a false match. In this work, 
the ratio is set to $0.6$, i.e. when 

\begin{equation}
\frac{MD_0}{MD_1}>0.6,
\label{eq7}
\end{equation}
the match will be considered a false match and the matching point pair ($F_0$ and $F_1$) will also be discarded.

\begin{figure}[ht]
\centering
\includegraphics[width=0.48\textwidth]{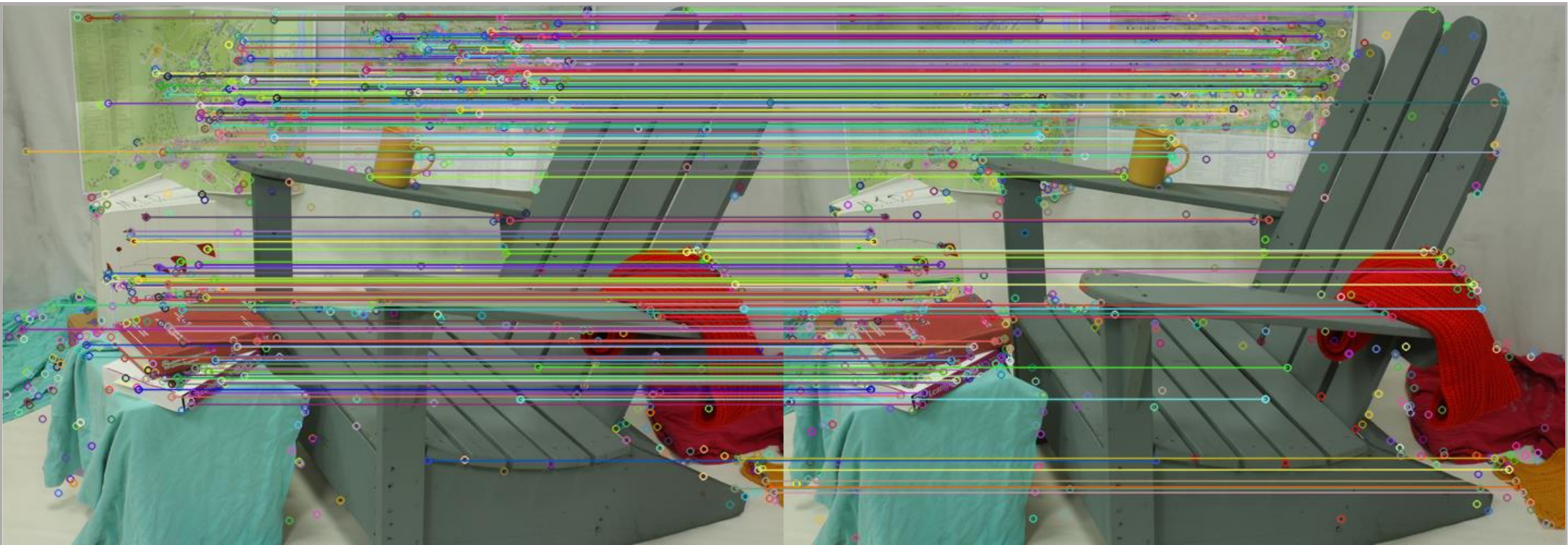}
\caption{SIFT Matching Pairs}
\label{SIFTMatching}
\end{figure}

A visualization of matching result is shown in the Fig. \ref{SIFTMatching}, which is implemented by OpenCV. Where the circles represent the feature points that were detected and the lines connecting the left and right view images represent a matching pair of feature points. It can be seen that not all feature points have a corresponding point.

\subsection{Feature Disparity Propagation}

In this section, we describe the basic idea of our method, feature disparity propagation, as the main contribution of this paper. The basic idea is that it allows the pixels $p$ with feature disparity to be propagated to the pixels $q$ with the high approximate at the same disparity. (Note that this procedure does not directly confirm the disparity value $d$, but increases the probability that the $d_q$ = $d_p$ at that pixel.)
feature disparity means the disparity of the pixel in the sparse disparity map, because the disparity in the sparse disparity map is obtained by matching pairs of feature points. We now explain details for the process of this step.

The propagation occurs in the cost matrix $CM$. We first define the sparse disparity map generated in the previous section as the set $S_{SDM}$, where all the pixels in the sparse disparity map are contained in the set $S_{SDM}$, namely $\forall p \in S_{SDM}$. In this work, $W_p$ is simply defined as a square window centred at pixel $p$, but $W_p$ could also be extended to use superpixels \cite{ref24}, \cite{ref25}, \cite{ref26}, such as a top ranked method NOSS\_ROB \cite{ref28}. The colour domain of the input image pair $IP$ is used to calculate $w(p,q)$.

\pagestyle{empty}
\begin{algorithm}[ht]  
      \caption{FEATURE DISPARITY PROPAGATION}  
      \label{algo1}  
      \begin{algorithmic}[1]  
        \Require  
          current \textit{$CM$}, \textit{$S_{SDM}$}, image pair \textit{$IP$}, propagating window \textit{$W_p$}
        \Ensure  
          updated \textit{$CM$}  
        \State \textbf{foreach}  pixel \textit{$p$} $\in$ {$S_{SDM}$} \textbf{do} 
        \State \textbf{if}{ \textit{$d_p$} $>$ 0} \textbf{ then}
        \State \textit{$CM(p,d_p)$} $\gets$ 0
        \State \textbf{foreach}  pixel \textit{$q$} $\in$ \textit{$W_p$} \textbf{do} 
        \State calculate \textit{ $w(p, q)$} according to Eq.\ref{eq7}
        \State \textit{$CM(q,d_p)$} $\gets$\textit{ $CM(q,d_p)$ } $\times$ $(1- 
w(p, q))$
        \State \textbf{end foreach}
        \State \textbf{end if}
        \State \textbf{end foreach}
      \end{algorithmic}  
    \end{algorithm}
    
We use this feature disparity propagation as shown in Algorithm 1. Initially, the pixels that has feature disparity in the sparse disparity map are found, i.e. the pixel $p$'s disparity in the sparse disparity map $d_p > 0$. Then, the matching cost of pixel $p$ under disparity $d_p$ is reduced to 0, i.e. $CM(p,d_p) = 0$. The purpose is to ensure that the matching cost of pixel $p$ under disparity $d$ is minimized.

\begin{figure}[ht]
\centering
\includegraphics[width=0.35\textwidth]{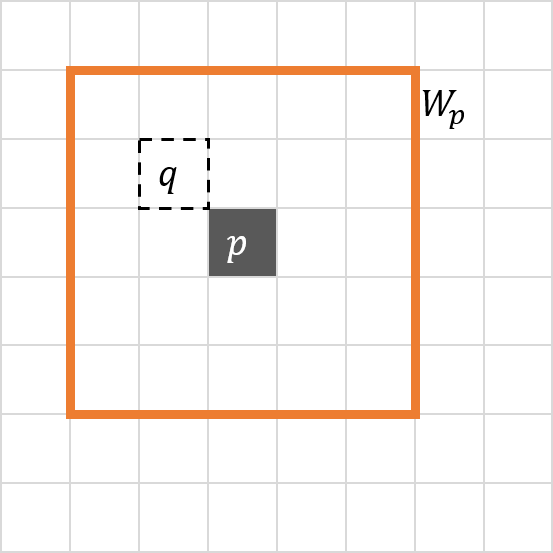}
\caption{Region of Window centred at $p$}
\label{window}
\end{figure}

Starting with the statement in line 4, we interfere with the matching cost of pixels $q$ neighbouring pixel $p$. The size of the neighbourhood is determined by $W_p$, as shown in Fig. \ref{window}. 

The similarity of all neighbouring pixels $q$ in $W_p$ to the centroid $p$ will be calculated separately. Here, $w(p,q)$ is a contrast-sensitive weight proposed in \cite{ref27} and implements the adaptive support weight idea. In this work, the weight function $w(p,q)$ shown as Equ.\ref{eq8} is used to compute the similarity between $p$ and $q$ by the pixel's colours.

\begin{equation}
w(p, q)=e^{-\left\|I_{L}(p)-I_{L}(q)\right\|_{1} / \gamma}
\label{eq8}\end{equation}

If there is a high similarity between p and q, the $w(p,q)$ function will return a high value. Meanwhile, $w(p,q) \in (0, 1]$. This idea is based on PatchMatch \cite{ref4}, \cite{ref23}, where when two pixels are of similar colour, there is a higher possibility that the two pixels are under the same disparity plane. 
We then "reward" the matching cost of the pixel q under disparity $d_p$ according to its colour similarity: The matching cost of point $q$ at $d_p$ is reduced proportionally, the higher the colour similarity to pixel $p$, the more the matching cost of point $q$ at $d_p$ will be reduced, as shown in line 6.

\begin{figure*}[htb]
\centering
\includegraphics[width=0.9\textwidth]{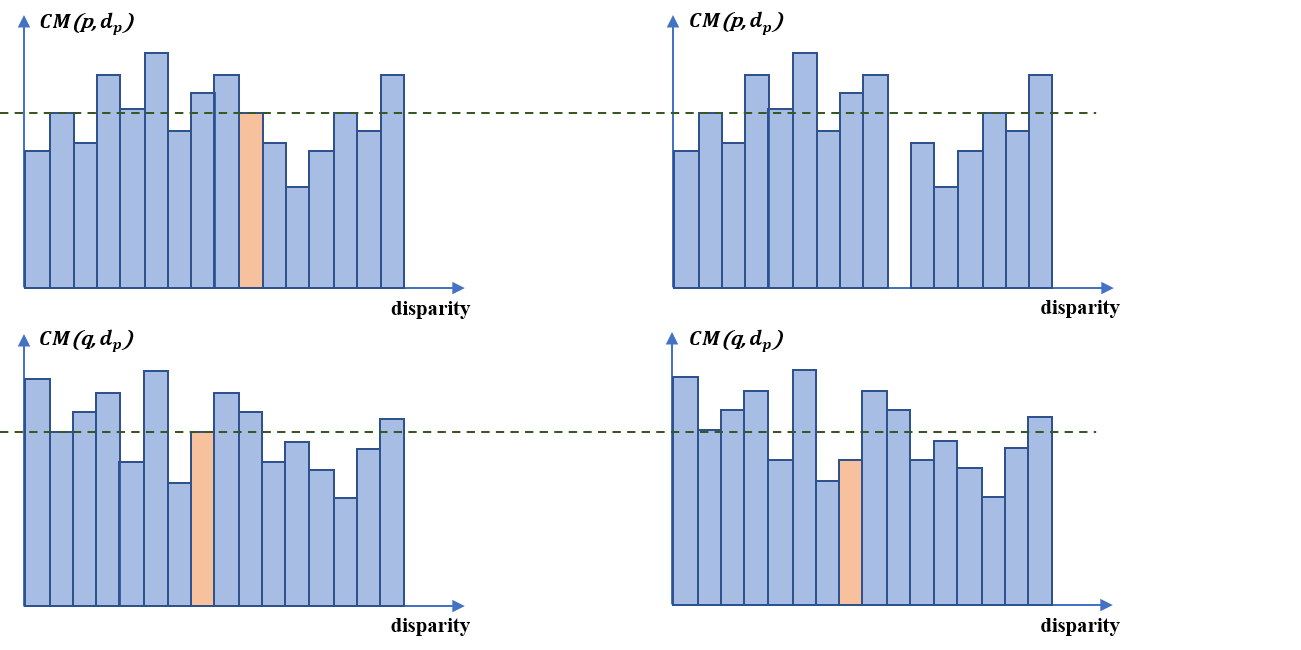}
\caption{Visualization of change of $CM$}
\label{propagation_cost_value}
\end{figure*}

The procedure for line 3 and line 6 is visualized in Fig. \ref{propagation_cost_value}. Since pixel $p$ has feature disparity, we consider that $d_p$ has a high possibility to be the disparity value of pixel $p$, therefore $CM(p, d_p)$ (i.e., the matching cost of pixel $p$ under disparity $d_p$) is decreased to 0. 
And for pixels $q$ in the neighbourhood of pixel $p$, we determine the amount of $CM(q, d_p)$ reduction by calculating the probability that they are in the same disparity plane. in this work, we simply use the colour similarity $w(p,q)$ to measure this probability, and indeed the probability can also be measured in combination with the distance between $p$ and $q$. This idea is applied in \cite{ref19}.

After Algo.\ref{algo1} is executed, the $CM$ is updated. In the following, $CM$ is represented as updated $CM$.

\subsection{Optimization and Post-processing}
The last step of our method is optimization and post-processing to generate the final dense disparity map from cost matrix $CM$. Inspired by method LocalExp's initialization using matching cost by MC-CNN, we use $CM$ instead the matching cost by MC-CNN for initialization of LocalExp. Meanwhile, LocalExp based on alpha-expansion in Graph-Cuts \cite{GC}, \cite{AlphaExpansionInGC} allows for better propagation of feature disparity in the dense disparity map.

\subsection{Experiments}
Our method is evaluated on the Middlebury stereo benchmark. Limited by the computation resource, our method was evaluated using images in quarter resolution, therefore we especially compared it with other methods that run in same resolution. It includes the state-of-the-art methods under each error threshold, such as ADSG \cite{ADSG}, ACR-GIF-OW \cite{ACR-GIF-OW}, FASW \cite{FASW} and SGM \cite{SGM}.

The user-defined parameters of Equ. \ref{eq3} are set as $\{ n, w_1,\\ w_2, w_3\} = \{ 3, 0.7, 0.2, 0.1\}$. $\alpha$ using to balance the influence of color and gradient term in Equ. \ref{eq4} is set to 0.8. For normalization of $\rho$ cost and Census cost in Equ. \ref{eq5}, we use $\{ \lambda{\rho}, \lambda_{census}\} = \{10, 30\}$. For Algo. \ref{algo1}, the size of feature propagation windows $W_p$ is set to $5 \times 5$ pixels, and $\gamma$ of the weight function Equ. \ref{eq8} for similarity calculating is defined as 10.

Our experiments are conducted on a laptop platform equipped with an Intel Core i3-8130U CPU (2.20 GHz $\times$ 2 physical cores) and with 4 GB memory. All methods are programming using C++ and OpenCV \cite{ref22}.

Our proposed cost matrix construction methods are only using a CPU implemented. We accelerate the calculation of cost matrix construction by parallel computing using 2 CPU cores. The parallel computing is implemented based on OpenMP \cite{ref29}, which is a specific library for parallel programming.

\subsection{Benchmark Result}
\begin{figure*}[htb]
\centering
\includegraphics[width=0.9\textwidth]{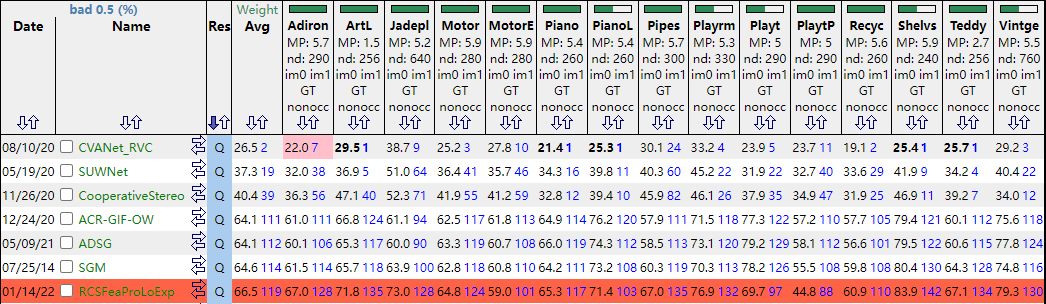}
\caption{Middlebury Benchmark V3 for the bad 0.5 nonocc Metric}
\label{result}
\end{figure*}

Our method RCSFeaProLoExp has been tested in the trainning set only so far, and the results are presented in Fig. \ref{result}. According to the diagram, the effect of method ACR-GIF-OW \cite{ACR-GIF-OW}, ADSG \cite{ADSG}, SGM \cite{SGM} and ours are in the second tier, with a rating of about 65. Visually, compared to their method, the edges in our result dispariry map are sharper.

\section{Conclusions}
In this paper, we propose a method that can using feature matching to assist stereo matching: feature disparity propagation. The core idea is to utilize a sparse disparity map to assist in building a dense disparity map, so that the dense disparity map can be obtained while retaining the more reliable disparity obtained by feature matching. Nevertheless, here we only use SIFT feature descriptors with KNN for demonstration, so it is still potential to improve the performance of this method by using better feature extraction, matching methods (e.g., SuperPixel \cite{ref6}, SuperGlue \cite{ref3} and other latest works).
Furthermore, we also propose a new matching cost measure, which we call $rho$-Census. It incorporates gradient and multi-scale information on the basis of the classical Ad-Census.

As mentioned above, It is potential to apply the latest feature extraction and matching methods to improve the performance of the method. Also based on feature disparity propagation, we believe that the scheme can also be adapted to achieve a higher resolution and more accurate disparity map by combining the low-resolution disparity map (such as those obtained from TOF) with the image pair. 

\printcredits





\end{document}